\documentclass[10pt,twocolumn,letterpaper]{article}

\usepackage[pagenumbers]{cvpr}   

\usepackage{preamble}

\usepackage[accsupp]{axessibility} 
\usepackage[pagebackref,breaklinks,colorlinks]{hyperref}

\hypersetup{
    filecolor=black,
    urlcolor=gray,
    citecolor=black,
}

\usepackage[capitalize]{cleveref}
\crefname{section}{Sec.}{Secs.}
\Crefname{section}{Section}{Sections}
\Crefname{table}{Table}{Tables}
\crefname{table}{Tab.}{Tabs.}

\begin{document}

\title{Explaining models relating objects and privacy}

\author{Alessio Xompero$^1$, Myriam Bontonou$^{2}$, Jean-Michel Arbona$^{2}$, Emmanouil Benetos$^1$, Andrea Cavallaro$^{1,3,4}$\\ 
$^1$Queen Mary University of London, United Kingdom, $^2$ENS de Lyon and CNRS, France,\\
$^3$Idiap Research Institute, $^4$École polytechnique fédérale de Lausanne, Switzerland\\
{\tt\small \{a.xompero,emmanouil.benetos\}@qmul.ac.uk}, 
{\tt\small \{myriam.bontonou,jeanmichel.arbona\}@ens-lyon.fr},\\{\tt\small a.cavallaro@idiap.ch}
}

\maketitle

\begin{abstract}
Accurately predicting whether an image is private before sharing it online is difficult due to the vast variety of content and the subjective nature of privacy itself. In this paper, we evaluate privacy models that use objects extracted from an image to determine why the image is predicted as private. To explain the decision of these models, we use feature-attribution to identify and quantify which objects (and which of their features) are more relevant to privacy classification with respect to a reference input (i.e., no objects localised in an image) predicted as public. We show that the presence of the person category and its cardinality is the main factor for the privacy decision. Therefore, these models mostly fail to identify private images depicting documents with sensitive data, vehicle ownership, and internet activity, or public images with people (e.g., an outdoor concert or people walking in a public space next to a famous landmark). As baselines for future benchmarks, we also devise two strategies that are based on the person presence and cardinality and achieve comparable classification performance of the privacy models.
\blfootnote{Alessio Xompero and Myriam Bontonou equally contributed. Myriam Bontonou is also affiliated with Inserm, France, and Jean-Michel Arbona with Univ Lyon and LBMC Lyon, France.}
\end{abstract}

\section{Introduction}

People take photos in a large variety of situations (e.g., at a party, of themselves, of a landmark, or of friends, family, animals, or food) and share them on social media platforms, often lacking awareness of privacy risks associated with their sharing~\cite{Ferrarello2022DRS,Cabarcos2023PoPETs}. Images may contain a set of objects that reveal private information about a person or be associated with a specific location or event that the person is attending. Therefore, an automatic warning prior to sharing could help users protect their privacy~\cite{Zerr2012SIGIR,Orekondy2017ICCV,Stoidis2022BigMM}.

\begin{figure*}[t!]
    \centering
    \includegraphics[width=0.9\linewidth]{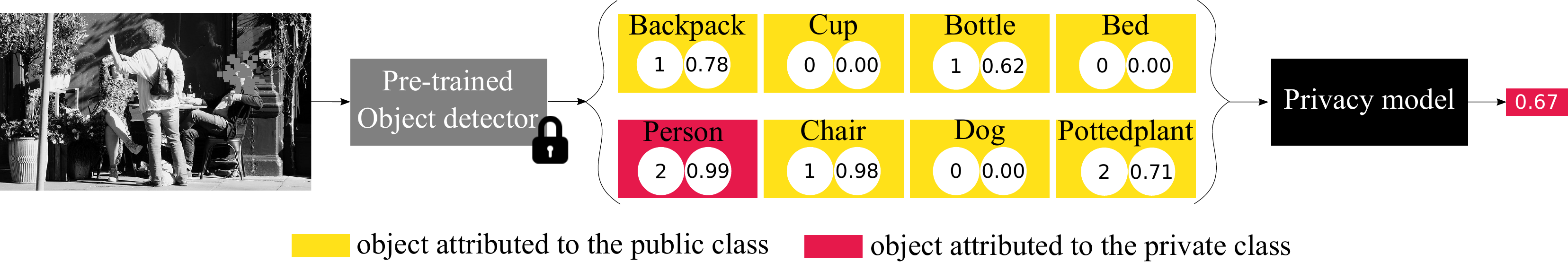}
    \caption{Two-stage privacy method: a pre-trained object detector identifies concepts (e.g., objects, scene type) within an image and a privacy model is trained to classify an image as private or public, considering the cardinality and confidence level of the extracted objects (numbers below each object).  
    The input image is from the PrivacyAlert dataset~\cite{Zhao2022ICWSM_PrivacyAlert}, with obfuscation added on the face of the person.}
    \label{fig:teaser}
    \vspace{-13pt}
\end{figure*}

Privacy classification methods are trained on datasets annotated by one or multiple annotators with a binary label (public or private)~\cite{Zerr2012SIGIR,Zhao2022ICWSM_PrivacyAlert,Yang2020PR}.
As the notion of privacy varies among people and also depends on the context, the annotation in these datasets is potentially ambiguous.
Most of the existing works design methods that aimed at improving the classification performance on these datasets. We categorise existing methods for image privacy as single-stage and two-stage. \textit{Single-stage} methods directly train or fine-tune a deep neural network (DNN) from the images~\cite{Orekondy2017ICCV}. \textit{Two-stage} methods uses DNNs (e.g., convolutional neural networks or CNNs) to extract concepts (i.e., objects, scenes) from the images followed by a privacy classifier in the second-stage, such as a Multi-Layer Perceptron (MLP) or a graph neural network (GNN)~\cite{Tonge2016AAAI,Tonge2018AAAI,Baranouskaya2023ICIP,Stoidis2022BigMM,Yang2020PR}. Two-stage methods can be further split into end-to-end training or hybrid. \textit{End-to-end training} based methods fine-tune the DNNs to initialise the concept features for the privacy classifier~\cite{Stoidis2022BigMM,Yang2020PR}.
\textit{Hybrid} methods extract concepts from the images with a pre-trained detector or multi-label image classifier~\cite{Tonge2016AAAI,Tonge2018AAAI,Baranouskaya2023ICIP}.  

In this paper, we explain the decisions made by a range of privacy classifiers that use as input the cardinality and confidence features of objects identified in an image (see Fig.~\ref{fig:teaser}). Among many existing explainability methods~\cite{Ribeiro2016KDD_LIME,Ribeiro2016ICMLw_LIME,sundararajan2017axiomatic,Petsiuk2018BMVC_rise,Yuan2023TPAMI_Survey}, we select integrated gradients~\cite{sundararajan2017axiomatic} that is computationally efficient and attributes the decision of the privacy models to the identified objects and their features with respect to a reference input. This reference input consists of features with zero values to represent the case of no objects localised in an image and hence classified as public. Based on the findings from the explainability analysis, we define two simple strategies using people presence as main driving factor to determine whether an image is private. These explainable-by-design strategies achieve comparable performance to the more complex privacy-decision models\footnote{Code: \url{https://github.com/graphnex/ig-privacy}}. As baselines in future comparisons, these strategies will also enable the design of explainable and more accurate privacy models that capture and use relationships between concepts beyond the only presence and cardinality of people in images.

\section{Problem formulation}

Let $I$ be an image and $f_{\theta}(\cdot)$ a privacy model trained on a dataset $\mathcal{D}=\{(I, y)_n\}_{n=1}^N$ to predict a class $y \in \{0,1\}$, where 0 denotes public and 1 private, $\theta$ contains the model parameters, and $N$ is the number of images in the training dataset. We consider the privacy model to map the outputs of other models to the predicted class $y$: $y=f_{\theta}(d_{\eta}(I))$, where $\eta$ contains pre-trained parameters. For example, $d_{\eta}(\cdot)$ can be a pre-trained object detector that localises a set of objects with their confidence in the image $I$. We refer to the pre-defined categories outputted by these pre-trained models as concepts.
Therefore, let $\mathcal{X} = \{\mathbf{x}^c | c=0, \ldots, C-1\}$ be the set of $C$ concepts with their $F$-dimensional feature vectors $\mathbf{x}^c = \left[ x^c_0, \ldots, x^c_{F-1} \right]$ that are provided as input to the privacy model. 

Our objective is to explain why the trained model $f_{\theta}(\cdot)$ predicts the label $y = 1$ for a given image $I$ (observable explanation~\cite{AlRegib2022SPM}). Specifically, we want to determine which concepts contribute to the prediction of the private label for the input image. To this end, we use post-hoc explainability to assign a score to each feature of each concept, \mbox{$\boldsymbol{\phi}(x_j^c) \in \{-1, 1\}$}.

Following previous works~\cite{Tonge2016AAAI,Yang2020PR,Stoidis2022BigMM}, we consider objects as concepts  and we use a pre-trained object detector to localise a pre-defined set of objects~\cite{Redmon2018YOLOv3} (i.e., $C=80$ for the COCO dataset~\cite{Lin2018ECCV_COCO}). We define two features ($F=2$) for each object: cardinality, $x^c_0 \in \mathbb{N}$ and confidence, $x_1^c \in [0,1]$. For cardinality, we count the number of instances localised in an image and belonging to each object. If no instances are localised for an image, then the cardinality is set to 0. For confidence, we retain the value of the most confident object instance if multiple instances of the same object are localised in an image. The privacy model could be an MLP or a GNN~\cite{Baranouskaya2023ICIP,Tonge2016AAAI,Tonge2018AAAI,Stoidis2022BigMM,Yang2020PR}. For MLP, the input is the concatenation of all object features, resulting in a vector of dimensionality $C F$: \mbox{$\mathbf{x} = [ \ldots, \mathbf{x}^c, \ldots], \forall \mathbf{x}^c \in \mathcal{X}$}. For GNN, the input is a $C\times F$ matrix of the object features, where each row corresponds to a node of a graph. For simplicity, we use the set $\mathcal{X}$ as input of the privacy model, independently of the representation: $f_{\theta}(\mathcal{X})$.

\section{Explaining image privacy predictions}

In this section, we describe the dataset and models used for image privacy (see the Supplementary Material document for additional details), and discuss their classification performance. We explain the models decision and analyse the explainability results. 

\noindent{\bf Dataset.} 
We use PrivacyAlert~\cite{Zhao2022ICWSM_PrivacyAlert} as a recent image privacy dataset $\mathcal{D}$ for our evaluations and analyses. PrivacyAlert has 6,800 images\footnote{PrivacyAlert provides links to images on Flickr whose license was falling under Public Domain~\cite{Zhao2022ICWSM_PrivacyAlert}.  Note that 7 images are no longer available and we re-train and evaluate models excluding these images.} split into a training set of 3,136 images (788 private images and 2,348 public images), a validation set of 1,864 images (466 private images and 1,398 public images), and a testing set of 1,800 images (450 private images and 1,350 public images), as originally described by the authors~\cite{Zhao2022ICWSM_PrivacyAlert}. The dataset has a high class imbalance towards the public images (ratio of about 3:1).

\begin{figure*}[t!]
    \centering
\begin{minipage}{0.18\textwidth}
\centering
~
\includegraphics[height=0.5\linewidth]{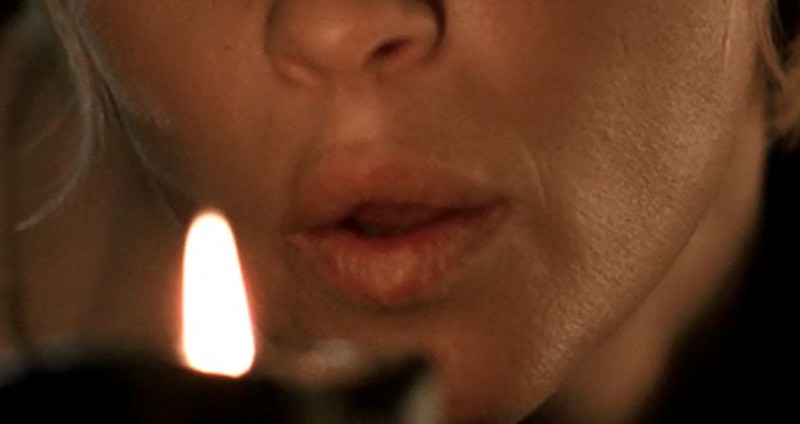}
\\
\begin{tikzpicture}
\begin{axis}[
    width=3.6cm,
    height=2.5cm,
    axis lines=left,
    ybar,
    ymin=0,
    ymax=2.2,
    ylabel style={align=center},
    yticklabel style={text width=0.42cm, align=right},
    xlabel style={align=center},
    xlabel={},
    ylabel={\footnotesize Features},
    ytick={0,1,2},
    ylabel style={yshift=-0.2cm},
    tick label style={font=\scriptsize},
    symbolic x coords={A, B},
    xticklabels={}, 
    xtick=data,
    bar width=4pt,
    enlarge x limits={0.2},
    ]
  \addplot[fill=blue!40, draw=none] coordinates {(A,1)};
  \addplot[fill=blue!20, draw=none] coordinates {(A,1)};
\end{axis}
\end{tikzpicture}
\\
\begin{tikzpicture}
\begin{axis}[
    width=3.6cm,
    height=2.5cm,
    axis lines=left,
    ybar,
    ymin=-0.2,
    ymax=0.6,
    ylabel style={align=center},
    xlabel style={align=center},
    xlabel={},
    ylabel={\footnotesize IG score},
    ytick={0,0.5},
    ylabel style={yshift=-1ex},
    symbolic x coords={A},
    axis x line shift=-0.2,
    xticklabels={Person}, 
    tick label style={font=\scriptsize},
    xtick=data,
    bar width=4pt,
    enlarge x limits={0.2},
    ]
  \addplot[fill=purple!60, draw=none] coordinates {(A,0.33)};
  \addplot[fill=pink!60, draw=none] coordinates {(A,0.24)};
\end{axis}
\end{tikzpicture}
\end{minipage}
\begin{minipage}{0.18\textwidth}
\centering
\includegraphics[height=0.5\linewidth]{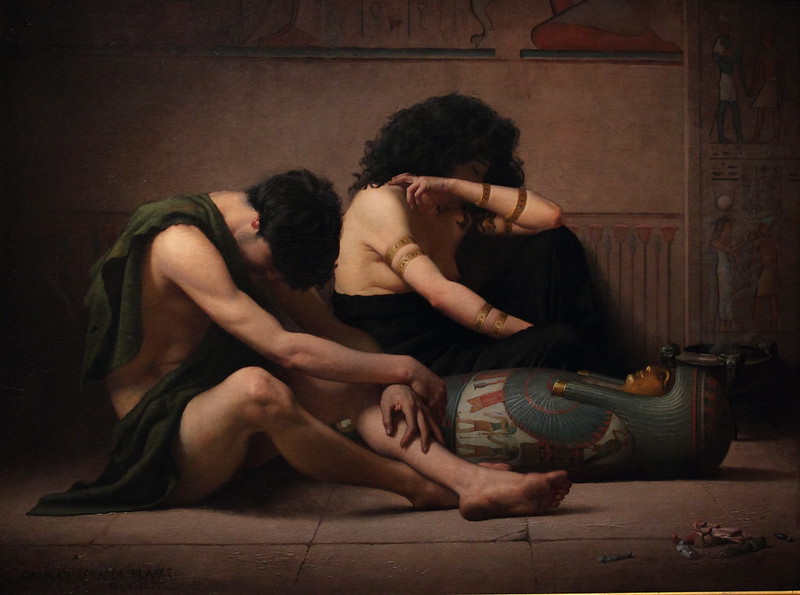}\\
\begin{tikzpicture}
\begin{axis}[
    width=3.6cm,
    height=2.5cm,
    axis lines=left,
    ybar,
    ymin=0,
    ymax=2.2,
    ylabel style={align=center},
    yticklabel style={text width=0.42cm, align=right},
    xlabel style={align=center},
    xlabel={},
    ylabel={},
    ytick={0,1,2},
    ylabel style={yshift=-1ex},
    symbolic x coords={A, B},
    xticklabels={}, 
    tick label style={font=\scriptsize},
    xtick=data,
    bar width=4pt,
    enlarge x limits={0.2},
    ]
  \addplot[fill=blue!40, draw=none] coordinates {(A,1)};
  \addplot[fill=blue!20, draw=none] coordinates {(A,2)};
\end{axis}
\end{tikzpicture}
\\
\begin{tikzpicture}
\begin{axis}[
    width=3.6cm,
    height=2.5cm,
    axis lines=left,
    ybar,
    ymin=-0.2,
    ymax=0.6,
    ylabel style={align=center},
    xlabel style={align=center},
    xlabel={},
    ylabel={},
    ytick={0,0.5},
    ylabel style={yshift=-1ex},
    symbolic x coords={A, B},
    axis x line shift=-0.2,
    xticklabels={Person}, 
    tick label style={font=\scriptsize},
    xtick=data,
    bar width=4pt,
    enlarge x limits={0.2},
    ]
  \addplot[fill=purple!60, draw=none] coordinates {(A,0.39)};
  \addplot[fill=pink!60, draw=none] coordinates {(A,0.02)};
\end{axis}
\end{tikzpicture}
\end{minipage}
\begin{minipage}{0.18\textwidth}
\centering
\includegraphics[height=0.5\linewidth]{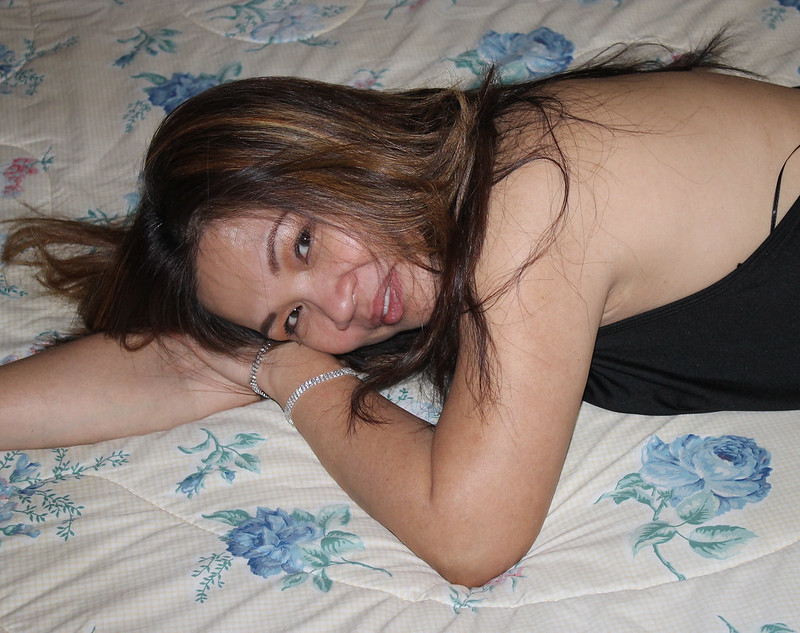}\\
\begin{tikzpicture}
\begin{axis}[
    width=3.6cm,
    height=2.5cm,
    axis lines=left,
    ybar,
    ymin=0,
    ymax=2.2,
    ylabel style={align=center},
    xlabel style={align=center},
    yticklabel style={text width=0.42cm, align=right},
    xlabel={},
    ylabel={},
    ytick={0,1,2},
    ylabel style={yshift=-1ex},
    symbolic x coords={A, B}, 
    xticklabels={}, 
    tick label style={font=\scriptsize},
    xtick=data,
    bar width=4pt,
    enlarge x limits={0.2},
    ]
  \addplot[fill=blue!40, draw=none] coordinates {(A,1) (B,1)};
  \addplot[fill=blue!20, draw=none] coordinates {(A,1) (B,1)};
\end{axis}
\end{tikzpicture}
\\
\begin{tikzpicture}
\begin{axis}[
    width=3.6cm,
    height=2.5cm,
    axis lines=left,
    ybar,
    ymin=-0.2,
    ymax=0.6,
    ylabel style={align=center},
    xlabel style={align=center},
    xlabel={},
    ylabel={},
    ytick={0,0.5},
    ylabel style={yshift=-1ex},
    symbolic x coords={A, B},
    axis x line shift=-0.2,
    xticklabels={Person,Bed},
    tick label style={font=\scriptsize},
    xtick=data,
    bar width=4pt,
    enlarge x limits={0.2},
    ]
  \addplot[fill=purple!60, draw=none] coordinates {(A,0.24) (B,0.16)};
  \addplot[fill=pink!60, draw=none] coordinates {(A,0.2) (B,0.16)};
\end{axis}
\end{tikzpicture}
\end{minipage}
\begin{minipage}{0.18\textwidth}
\centering
\includegraphics[height=0.5\linewidth]{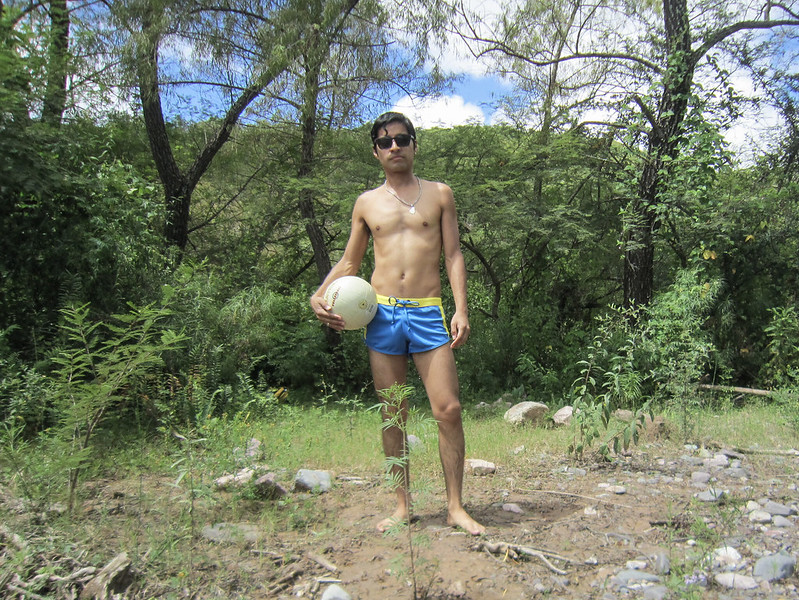}\\
\begin{tikzpicture}
\begin{axis}[
    width=3.6cm,
    height=2.5cm,
    axis lines=left,
    ybar,
    ymin=0,
    ymax=2.2,
    ylabel style={align=center},
    xlabel style={align=center},
    yticklabel style={text width=0.42cm, align=right},
    xlabel={},
    ylabel={},
    ytick={0,1,2},
    ylabel style={yshift=-1ex},
    symbolic x coords={A, B},
    xticklabels={}, 
    tick label style={font=\scriptsize},
    xtick=data,
    bar width=4pt,
    enlarge x limits={0.2},
    ]
  \addplot[fill=blue!40, draw=none] coordinates {(A,1) (B,0.99)};
  \addplot[fill=blue!20, draw=none] coordinates {(A,1) (B,1)};
\end{axis}
\end{tikzpicture}
\\
\begin{tikzpicture}
\begin{axis}[
    width=3.6cm,
    height=2.5cm,
    axis lines=left,
    ybar,
    ymin=-0.2,
    ymax=0.6,
    ylabel style={align=center},
    xlabel style={align=center},
    xlabel={},
    ylabel={},
    ytick={0,0.5},
    ylabel style={yshift=-1ex},
    symbolic x coords={A, B},
    axis x line shift=-0.2,
    xticklabels={Person,Frisbee}, 
    tick label style={font=\scriptsize},
    xtick=data,
    bar width=4pt,
    enlarge x limits={0.2},
    ]
  \addplot[fill=purple!60, draw=none] coordinates {(A,0.32) (B,0.16)};
  \addplot[fill=pink!60, draw=none] coordinates {(A,0.23) (B,-0.03)};
\end{axis}
\end{tikzpicture}
\end{minipage}
\begin{minipage}{0.18\textwidth}
\centering
\includegraphics[height=0.5\linewidth]{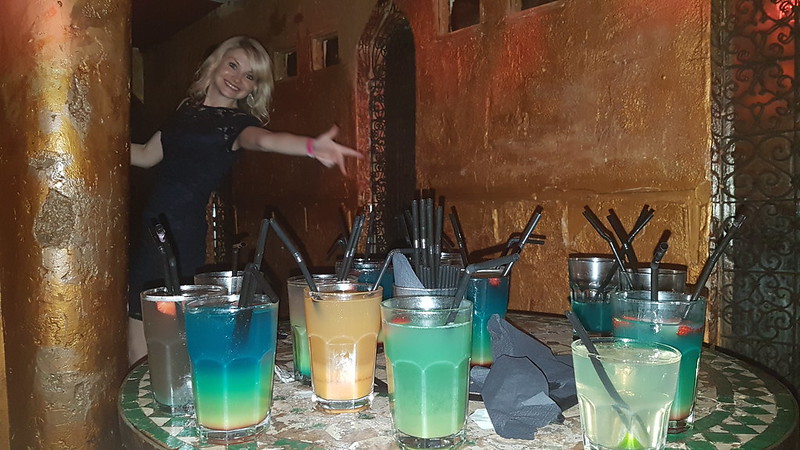}\\
\begin{tikzpicture}
\begin{axis}[
    width=3.6cm,
    height=2.5cm,
    axis lines=left,
    ybar,
    ymin=0,
    ymax=6.6,
    ylabel style={align=center},
    xlabel style={align=center},
    xlabel={},
    ylabel={},
    ytick={0,6},
    yticklabel style={text width=0.42cm, align=right},
    ylabel style={yshift=-1ex},
    symbolic x coords={A, B},
    xticklabels={}, 
    tick label style={font=\scriptsize},
    xtick=data,
    bar width=4pt,
    enlarge x limits={0.2},
    ]
  \addplot[fill=blue!40, draw=none] coordinates {(A,0.99) (B,0.99)};
  \addplot[fill=blue!20, draw=none] coordinates {(A,1) (B,6)};
\end{axis}
\end{tikzpicture}
\\
\begin{tikzpicture}
\begin{axis}[
    width=3.6cm,
    height=2.5cm,
    axis lines=left,
    ybar,
    ymin=-0.2,
    ymax=0.6,
    ylabel style={align=center},
    xlabel style={align=center},
    xlabel={},
    ylabel={},
    ytick={0,0.5},
    ylabel style={yshift=-1ex},
    symbolic x coords={A, B},
    axis x line shift=-0.2,
    xticklabels={Person,Cup}, 
    tick label style={font=\scriptsize},
    xtick=data,
    bar width=4pt,
    enlarge x limits={0.2},
    ]
  \addplot[fill=purple!60, draw=none] coordinates {(A,0.32) (B,0.03)};
  \addplot[fill=pink!60, draw=none] coordinates {(A,0.22) (B,-0.1)};
\end{axis}
\end{tikzpicture}
\end{minipage}

\begin{minipage}{0.18\textwidth}
\centering
\includegraphics[height=0.5\linewidth]{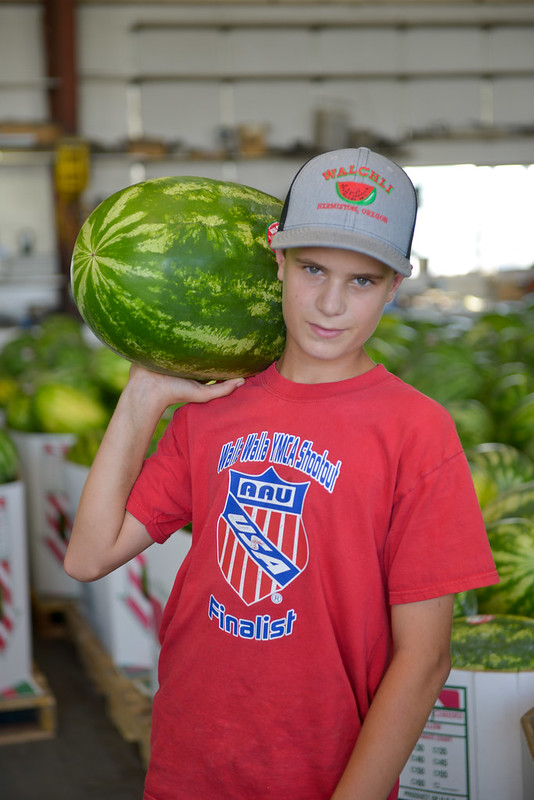}\\
\begin{tikzpicture}
\begin{axis}[
    width=3.6cm,
    height=2.5cm,
    axis lines=left,
    ybar,
    ymin=0,
    ymax=2.2,
    ylabel style={align=center},
    xlabel style={align=center},
    xlabel={},
    ylabel={Features},
    ytick={0,1,2},
    ylabel style={yshift=-0.2cm, font=\footnotesize},
    yticklabel style={text width=0.42cm, align=right},
    symbolic x coords={A},
    xticklabels={},
    tick label style={font=\scriptsize},
    xtick=data,
    bar width=4pt,
    enlarge x limits={0.2},
    ]
  \addplot[fill=blue!40, draw=none] coordinates {(A,1)};
  \addplot[fill=blue!20, draw=none] coordinates {(A,1)};
\end{axis}
\end{tikzpicture}
\\
\begin{tikzpicture}
\begin{axis}[
    width=3.6cm,
    height=2.5cm,
    axis lines=left,
    ybar,
    ymin=-0.2,
    ymax=0.6,
    ylabel style={align=center},
    xlabel style={align=center},
    xlabel={},
    ylabel={IG score},
    ytick={0,0.5},
    ylabel style={yshift=-1ex, font=\footnotesize},
    symbolic x coords={A},
    axis x line shift=-0.2,
    xticklabels={Person}, 
    tick label style={font=\scriptsize},
    x tick label style={yshift=-1ex},
    xtick=data,
    bar width=4pt,
    enlarge x limits={0.2},
    ]
  \addplot[fill=purple!60, draw=none] coordinates {(A,0.33)};
  \addplot[fill=pink!60, draw=none] coordinates {(A,0.23)};
\end{axis}
\end{tikzpicture}
\end{minipage}
\begin{minipage}{0.18\textwidth}
\centering
\includegraphics[height=0.5\linewidth]{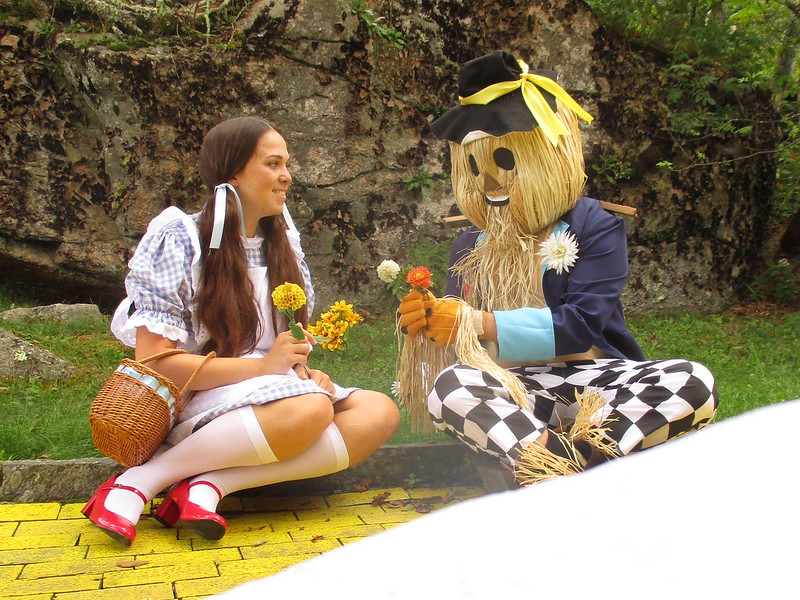}\\
\begin{tikzpicture}
\begin{axis}[
    width=3.6cm,
    height=2.5cm,
    axis lines=left,
    ybar,
    ymin=0,
    ymax=2.2,
    ylabel style={align=center},
    xlabel style={align=center},
    xlabel={},
    ylabel={},
    ytick={0,1,2},
    yticklabel style={text width=0.42cm, align=right},
    ylabel style={yshift=-1ex},
    symbolic x coords={A,},
    xticklabels={},
    tick label style={font=\scriptsize},
    xtick=data,
    bar width=4pt,
    enlarge x limits={0.2},
    ]
  \addplot[fill=blue!40, draw=none] coordinates {(A,1)};
  \addplot[fill=blue!20, draw=none] coordinates {(A,2)};
\end{axis}
\end{tikzpicture}
\\
\begin{tikzpicture}
\begin{axis}[
    width=3.6cm,
    height=2.5cm,
    axis lines=left,
    ybar,
    ymin=-0.2,
    ymax=0.6,
    ylabel style={align=center},
    xlabel style={align=center},
    xlabel={},
    ylabel={},
    ytick={0,0.5},
    ylabel style={yshift=-1ex},
    symbolic x coords={A, B},
    axis x line shift=-0.2,
    xticklabels={Person}, 
    tick label style={font=\scriptsize},
    x tick label style={yshift=-1ex},
    xtick=data,
    bar width=4pt,
    enlarge x limits={0.2},
    ]
  \addplot[fill=purple!60, draw=none] coordinates {(A,0.39)};
  \addplot[fill=pink!60, draw=none] coordinates {(A,0.02)};
\end{axis}
\end{tikzpicture}
\end{minipage}
\begin{minipage}{0.18\textwidth}
\centering
\includegraphics[height=0.5\linewidth]{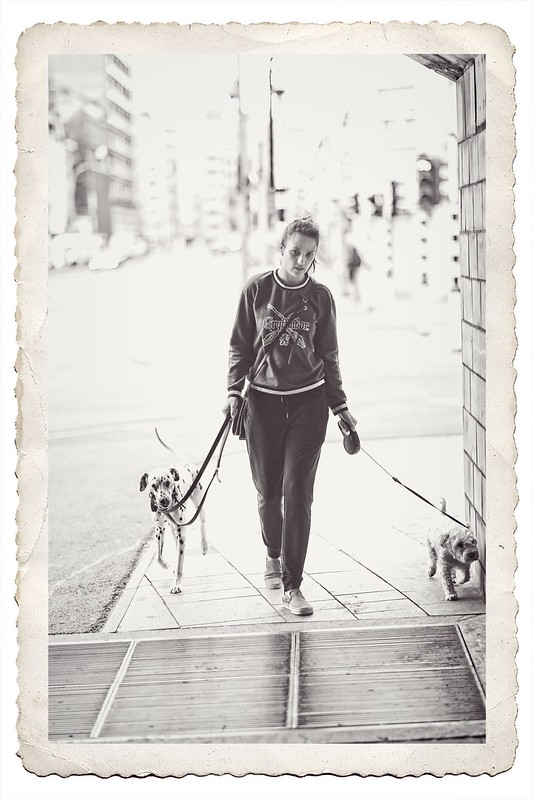}\\
\begin{tikzpicture}
\begin{axis}[
    width=3.6cm,
    height=2.5cm,
    axis lines=left,
    ybar,
    ymin=0,
    ymax=2.2,
    ylabel style={align=center},
    xlabel style={align=center},
    xlabel={},
    ylabel={},
    ytick={0,1,2},
    yticklabel style={text width=0.42cm, align=right},
    ylabel style={yshift=-1ex},
    symbolic x coords={A, B},
    xticklabels={},
    tick label style={font=\scriptsize},
    xtick=data,
    bar width=4pt,
    enlarge x limits={0.2},
    ]
  \addplot[fill=blue!40, draw=none] coordinates {(A,1) (B,0.84)};
  \addplot[fill=blue!20, draw=none] coordinates {(A,1) (B,2)};
\end{axis}
\end{tikzpicture}
\\
\begin{tikzpicture}
\begin{axis}[
    width=3.6cm,
    height=2.5cm,
    axis lines=left,
    ybar,
    ymin=-0.2,
    ymax=0.6,
    ylabel style={align=center},
    xlabel style={align=center},
    xlabel={},
    ylabel={},
    ytick={0,0.5},
    ylabel style={yshift=-1ex},
    symbolic x coords={A, B},
    axis x line shift=-0.2,
    xticklabels={Person,Dog}, 
    tick label style={font=\scriptsize},
    x tick label style={yshift=-1ex},
    xtick=data,
    bar width=4pt,
    enlarge x limits={0.2},
    ]
  \addplot[fill=purple!60, draw=none] coordinates {(A,0.31) (B,0.07)};
  \addplot[fill=pink!60, draw=none] coordinates {(A,0.21) (B,-0.15)};
\end{axis}
\end{tikzpicture}
\end{minipage}
\begin{minipage}{0.18\textwidth}
\centering
\includegraphics[height=0.5\linewidth]{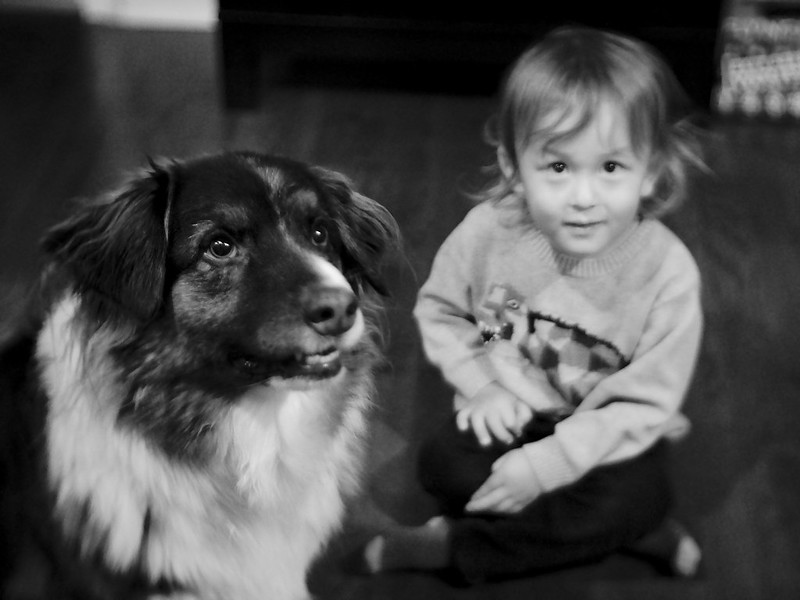}\\
\begin{tikzpicture}
\begin{axis}[
    width=3.6cm,
    height=2.5cm,
    axis lines=left,
    ybar,
    ymin=0,
    ymax=2.2,
    ylabel style={align=center},
    xlabel style={align=center},
    xlabel={},
    ylabel={},
    ytick={0,1,2},
    yticklabel style={text width=0.42cm, align=right},
    ylabel style={yshift=-1ex},
    symbolic x coords={A, B, C},
    xticklabels={},
    tick label style={font=\scriptsize},
    xtick=data,
    bar width=4pt,
    enlarge x limits={0.2},
    ]
  \addplot[fill=blue!40, draw=none] coordinates {(A,1) (B,1) (C,0.87)};
  \addplot[fill=blue!20, draw=none] coordinates {(A,1) (B,1) (C,1)};
\end{axis}
\end{tikzpicture}
\\
\begin{tikzpicture}
\begin{axis}[
    width=3.6cm,
    height=2.5cm,
    axis lines=left,
    ybar,
    ymin=-0.2,
    ymax=0.6,
    ylabel style={align=center},
    xlabel style={align=center},
    xlabel={},
    ylabel={},
    ytick={0,0.5},
    ylabel style={yshift=-1ex},
    symbolic x coords={A, B, C},
    axis x line shift=-0.2,
    xticklabels={Person,Dog,Sofa}, 
    tick label style={font=\scriptsize},
    x tick label style={yshift=-1ex},
    xtick=data,
    bar width=4pt,
    enlarge x limits={0.2},
    ]
  \addplot[fill=purple!60, draw=none] coordinates {(A,0.33) (B,0.11) (C,0.13)};
  \addplot[fill=pink!60, draw=none] coordinates {(A,0.24) (B,-0.17) (C,-0.06)};
\end{axis}
\end{tikzpicture}
\end{minipage}
\begin{minipage}{0.18\textwidth}
\centering
\includegraphics[height=0.5\linewidth]{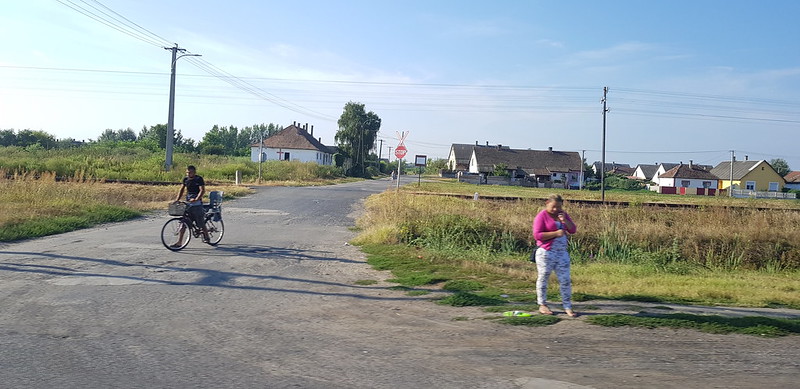}\\
\begin{tikzpicture}
\begin{axis}[
    width=3.6cm,
    height=2.5cm,
    axis lines=left,
    ybar,
    ymin=0,
    ymax=2.2,
    ylabel style={align=center},
    xlabel style={align=center},
    xlabel={},
    ylabel={},
    ytick={0,1,2},
    yticklabel style={text width=0.42cm, align=right},
    ylabel style={yshift=-1ex},
    symbolic x coords={A, B},
    xticklabels={},
    tick label style={font=\scriptsize},
    xtick=data,
    bar width=4pt,
    enlarge x limits={0.2},
    ]
  \addplot[fill=blue!40, draw=none] coordinates {(A,1) (B,0.98)};
  \addplot[fill=blue!20, draw=none] coordinates {(A,2) (B,1)};
\end{axis}
\end{tikzpicture}
\\
\begin{tikzpicture}
\begin{axis}[
    width=3.6cm,
    height=2.5cm,
    axis lines=left,
    ybar,
    ymin=-0.2,
    ymax=0.6,
    ylabel style={align=center},
    xlabel style={align=center},
    xlabel={},
    ylabel={},
    ytick={0,0.5},
    ylabel style={yshift=-1ex},
    symbolic x coords={A, B},
    axis x line shift=-0.2,
    xticklabels={Person,Bicycle}, 
    tick label style={font=\scriptsize},
    x tick label style={yshift=-1ex},
    xtick=data,
    bar width=4pt,
    enlarge x limits={0.2},
    ]
  \addplot[fill=purple!60, draw=none] coordinates {(A,0.39) (B,0.08)};
  \addplot[fill=pink!60, draw=none] coordinates {(A,0.02) (B,-0.08)};
\end{axis}
\end{tikzpicture}
\end{minipage}
 \caption{Sample of training images from PrivacyAlert~\cite{Zhao2022ICWSM_PrivacyAlert}  correctly predicted as private (first row) and incorrectly predicted as private (fourth row) by the graph-agnostic baseline~\cite{Dwivedi2023JMLR}, with their extracted object and features (blue bar plots) and the explanation scores (red bar plots) of Integrated Gradients (IG)~\cite{sundararajan2017axiomatic}. Darker colours (left bar) are associated with the confidence feature and lighter colours (right bar) with cardinality.  Positive IG scores support privacy, whereas negative IG scores support the public decision. 
 Note the different maximum limit for the y-axis in the top-right blue bar plot (fifth column, second row).
    }
    \vspace{-7pt}
    \label{fig:explainabilitysamples}
\end{figure*}
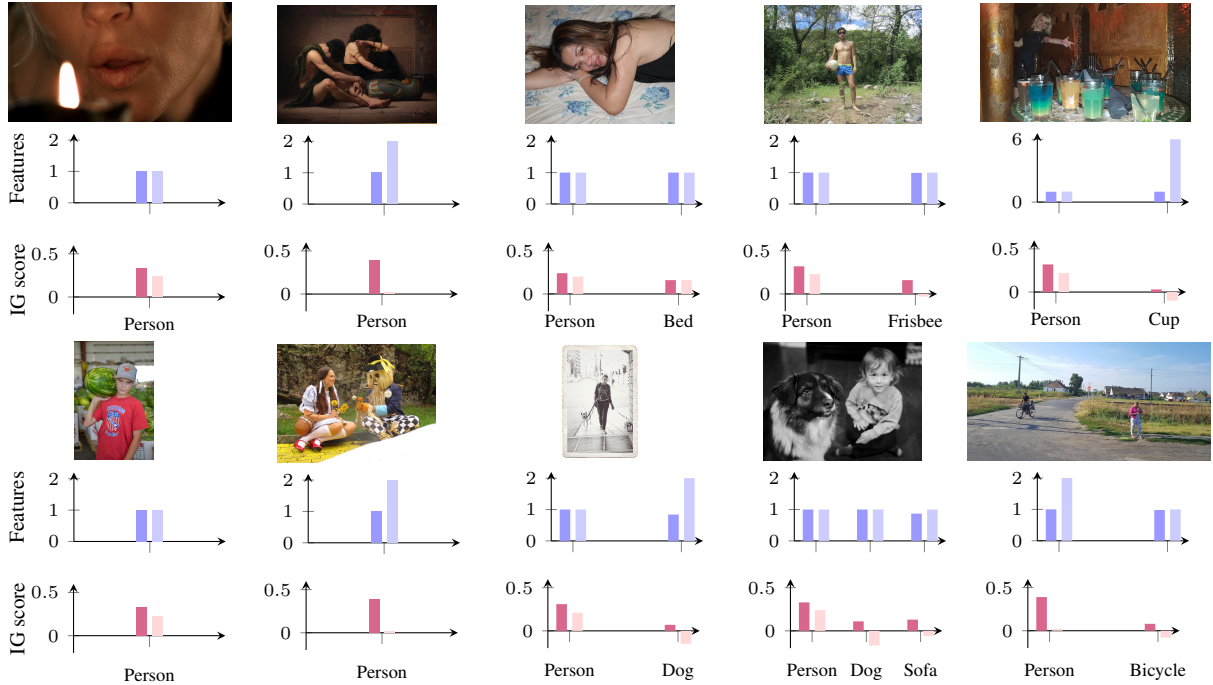

\noindent{\bf Methods.} 
We consider  MLP~\cite{Tonge2016AAAI,Baranouskaya2023ICIP}, two  graph-based models, GIP~\cite{Yang2020PR} and GPA~\cite{Stoidis2022BigMM}, and a graph-agnostic model (GA-MLP)~\cite{Dwivedi2023JMLR}.
The MLP aims at reproducing Tonge et al.'s method~\cite{Tonge2016AAAI} that uses Support Vector Machine as a privacy classifier and, as input, a binary feature vector of the top-$k$ most confident classes recognised by a pre-trained CNN for multi-label object recognition. In our case, we replace the multi-label classifier with the object detector and the object presence with the cardinality and confidence features of the identified objects. The MLP consists of 3 hidden layers, each with a 16-dimensionality hidden status and followed by batch normalisation. 
GIP and GPA modelled graphs to relate the objects with two privacy classes or the objects with each other, respectively, using a Graph Reasoning Model~\cite{Wang2018IJCAI} as GNN. These two models belong to the two-stage end-to-end training category. They fine-tune the CNNs in the first stage to initialise the node features and the CNN thus contribute to the privacy decision of the models. 
For a fair comparison, we adapt GIP and GPA to the two-stage hybrid approach by decoupling the GNN from the CNNs. We use only the GNN with the graph modelled by each method and with the cardinality and confidence as input node features. This adaptation allows us to assess the impact of the GNN as privacy model.
GA-MLP aims to replicate the steps of a GNN but without the graph structure (graph-agnostic)~\cite{Dwivedi2023JMLR}. 
To enable the training of the model, we independently project each node feature to a higher dimensionality vector with a fully connected layer shared among the nodes, and we then concatenate the projected features. Similarly to the multiple layers of a GNN, we refine the projected node features using three blocks, each consisting of a fully connected layer, a batch normalization layer~\cite{Ioffe2015ICML_BatchNorm}, a ReLU activation function, and a dropout layer~\cite{Srivastava2014DropoutAS}. 
We aggregate the refined features using global sum pooling and we provide the resulting global feature vector as input to an MLP-based classifier.

\noindent{\bf Classification.} Table~\ref{tab:classres} compares the classification performance of the privacy models on PrivacyAlert. Given the class imbalance of the dataset, we discuss the results in terms of recall on the private class and balanced accuracy (average recall of the two classes), reported as percentages. Both MLP and GA-MLP achieve a balanced accuracy of 71.60\% and 74.30\%, respectively, and an overall precision of 69.90\% and 70.20\%. GA-MLP correctly identifies more private images than MLP (higher recall in the private class). GPA and GIP models degenerate to predict (almost) all images as public, showing that the decoupled graph component based only on object features is not useful for privacy. This suggests that the models trained in the corresponding papers~\cite{Yang2020PR,Stoidis2022BigMM} were driven by the fine-tuning of CNNs.

\pgfplotstableread{beeswarm_gamlp_card.txt}\gamlpcard
\pgfplotstableread{beeswarm_gamlp_conf.txt}\gamlpconf
\pgfplotstableread{beeswarm_mlp_card.txt}\mlpcard
\pgfplotstableread{beeswarm_mlp_conf.txt}\mlpconf
\begin{figure*}[t!]
    \centering
    \begin{tikzpicture}
    \begin{axis}[
        width=.45\linewidth,
        y=0.42cm,
        xmin=-0.4,
        xmax=0.5,
        ymin=-0.5,
        ymax=4.5,
        ytick={0,1,2,3,4},
        yticklabels={person, chair, tie, bed, diningtable},
        label style={font=\footnotesize},
        tick label style={font=\scriptsize}, 
        ymajorgrids=true,
        y dir=reverse,
        title={\textbf{GA-MLP}},
        title style={font=\footnotesize},
        colormap/bluered,
            point meta=explicit
        ]
        \addplot+[scatter, only marks, mark size=1pt]table[x=IGs,y=concept, meta=card]\gamlpcard;
    \end{axis}
    \end{tikzpicture}
    \begin{tikzpicture}
    \begin{axis}[
        width=.45\linewidth,
        y=0.42cm,
        xmin=-0.4,
        xmax=0.5,
        ymin=-0.5,
        ymax=4.5,
        ytick={0,1,2,3,4},
        yticklabels={},
        label style={font=\footnotesize},
        tick label style={font=\scriptsize}, 
        ymajorgrids=true,
        y dir=reverse,
        title={\textbf{MLP} },
        title style={font=\footnotesize},
        colormap/bluered,
        colorbar,
        colorbar style={
                width=.02\columnwidth, 
                ytick={0, 1, 2, 3},
                yticklabels={0, 1, 2, $\geq 3$},
                yticklabel style={
                    tick label style={font=\scriptsize},
                },
                scaled y ticks=false,
                ylabel style={font=\scriptsize},
                ylabel={Concept cardinality},
            },
            point meta=explicit
        ]
        \addplot+[scatter, only marks, mark size=1pt]table[x=IGs,y=concept, meta=card]\mlpcard;
    \end{axis}
    \end{tikzpicture}
    \begin{tikzpicture}
    \begin{axis}[
        width=.45\linewidth,
        y=0.42cm,
        xmin=-0.4,
        xmax=0.5,
        ymin=-0.5,
        ymax=4.5,
        ytick={0,1,2,3,4},
        yticklabels={person, chair, tie, bed, diningtable},
        label style={font=\footnotesize},
        tick label style={font=\scriptsize}, 
        ymajorgrids=true,
        y dir=reverse,
        xlabel={Integrated Gradients score},
        title style={font=\footnotesize},
        colormap/bluered,
        colorbar style={
                width=.02\columnwidth, 
                yticklabel style={
                    tick label style={font=\scriptsize},
                },
                ytick={0, 1},
                scaled y ticks=false,
            },
            point meta=explicit
        ]
        \addplot+[scatter, only marks, mark size=1pt]table[x=IGs,y=concept, meta=conf]\gamlpconf;
        \addplot+[solid, gray, no marks]table[x=x,y=y,meta=z]{
        x    y    z
        0     -1  0
        0     6   1
        };
    \end{axis}
    \end{tikzpicture}
    \begin{tikzpicture}
    \begin{axis}[
        width=.45\linewidth,
        y=0.42cm,
        xmin=-0.4,
        xmax=0.5,
        ymin=-0.5,
        ymax=4.5,
        ytick={0,1,2,3,4},
        yticklabels={},
        label style={font=\footnotesize},
        tick label style={font=\scriptsize}, 
        ymajorgrids=true,
        y dir=reverse,
        xlabel={Integrated Gradients score},
        title style={font=\footnotesize},
        colormap/bluered,
        colorbar,
        colorbar style={
                width=.02\columnwidth, 
                yticklabel style={
                    tick label style={font=\scriptsize},
                },
                ytick={0, 1},
                scaled y ticks=false,
                ylabel={Concept confidence},
                ylabel style={font=\scriptsize,yshift=-2ex},
            },
            point meta=explicit
        ]
        \addplot+[scatter, only marks, mark size=1pt]table[x=IGs,y=concept, meta=conf]\mlpconf;
        \addplot+[solid, gray, no marks]table[x=x,y=y,meta=z]{
        x    y    z
        0     -1  0
        0     6   1
        };
    \end{axis}
    \end{tikzpicture}
    \caption{Comparison of the explainability scores across training images correctly classified as private by the graph-agnostic (GA-MLP) and MLP models on the training set of PrivacyAlert~\cite{Zhao2022ICWSM_PrivacyAlert}. We show only the top 5 objects based on the largest mean absolute explainability scores. Note that colours of the data points represent the value of the object feature. Also note the different limits of the colour bars. 
    }
    \label{fig:explconsmodels}
\end{figure*}
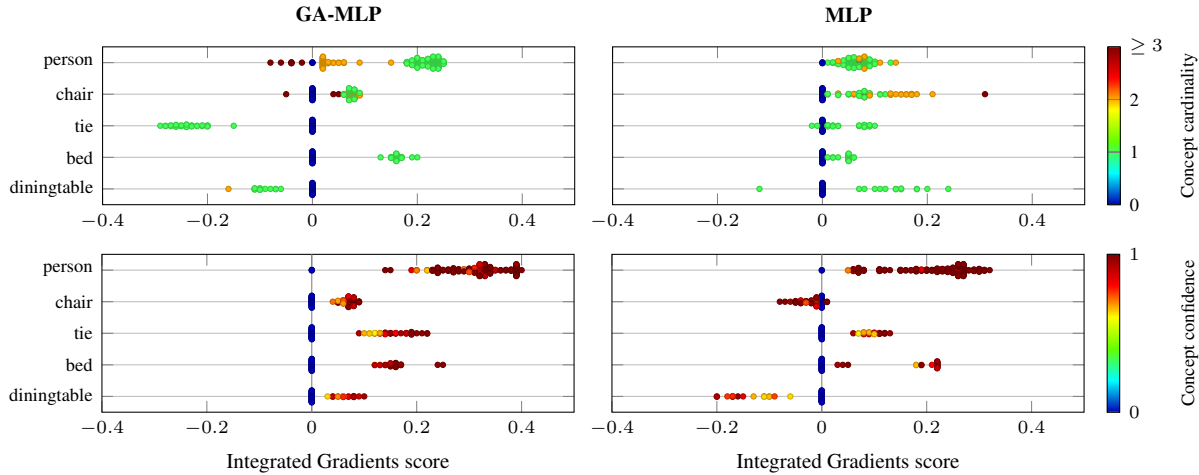

\noindent{\bf Explainability analysis.}
To explain the privacy models, we analyse the importance of the extracted concepts and their features to the decision of the models by using  Integrated Gradients (IG)~\cite{sundararajan2017axiomatic}. IG is a  post-hoc explainability method that is widely used to attribute the prediction of a model to the input features, resulting in a IG score per object feature $\boldsymbol{\phi}({x}_j^c)$. 
As IG is model-agnostic, the method can be applied to all gradient-based models. 
Specifically, IG compares the privacy prediction of a model $f_{\theta}$ for the input set of objects features $\mathcal{X}$ with the privacy prediction of the same model for a reference input \mbox{$\mathcal{R}=\{\mathbf{r}^c | c=0, \ldots, C-1\}$}. As private is the target class, we select a null-vector as our reference for each concept: $\mathbf{r}^c = \boldsymbol{0}, \forall c$ so that $f_\theta(\mathcal{R})=0$. This is equivalent to no objects detected in an image. 
IG also satisfies the \textit{completeness} axiom~\cite{sundararajan2017axiomatic},  
\begin{equation}
     f_\theta(\mathcal{X}) - f_\theta(\mathcal{R}) = \sum_{c=0}^{C-1}\sum_{j=0}^{F-1}\boldsymbol{\phi}({x}_j^c)\,,
\end{equation}
that quantifies the contribution of the features of all objects towards the decision of the model.  
Fig.~\ref{fig:explainabilitysamples} shows the explainabilty scores for a sample of images predicted as private by GA-MLP. Images are selected from the training set of PrivacyAlert based on the objects and their cardinality identified in the images, contrasting correct and incorrect predictions.
When images are correctly predicted as private (top row), high confidence in detecting a person significantly influences the decision of the model. On the contrary, the localisation of multiple individuals in an image tends to favour the public class. Public images are often misclassified due to the detection of \textit{person} (second row of images).
Fig.~\ref{fig:explconsmodels} compares the predictions and explainability of GA-MLP and MLP across all images correctly identified as private in the training set\footnote{The confidence to predict the reference input as private is 0.2 for MLP and 0.1 for GA-MLP.}. As for the previous analysis, \textit{person} is the most relevant concept for private predictions. Unlike MLP, GA-MLP favours the public class when three or more people are detected.

\begin{table}[t!]
    \centering
    \scriptsize
    \setlength\tabcolsep{7.5pt}
    \caption{Classification performance on the testing set of PrivacyAlert~\cite{Zhao2022ICWSM_PrivacyAlert}. All the models are using the same object detector to extract object features from the images. Note the failure of GPA~\cite{Stoidis2022BigMM} and GIP~\cite{Yang2020PR} adapted to the hybrid approach and using only GNN (recall of 100\% for public class, and precision and recall of 0\% for private class). Their original performance was driven by the dependence on CNNs~\cite{Stoidis2022BigMM,Yang2020PR}. 
    \vspace{-7pt}
    }
    \begin{tabular}{lrrrrrr}
    \toprule
    \multicolumn{1}{c}{\textbf{Method}} & \multicolumn{2}{c}{\textbf{Public}} & \multicolumn{2}{c}{\textbf{Private}} & \multicolumn{2}{c}{\textbf{Overall}}     \\
    \cmidrule(lr){2-3}\cmidrule(lr){4-5}\cmidrule(lr){6-7}
    & \multicolumn{1}{c}{P} & \multicolumn{1}{c}{R}  & \multicolumn{1}{c}{P} & \multicolumn{1}{c}{R}  & \multicolumn{1}{c}{P} & \multicolumn{1}{c}{BA} \\
    \midrule
    All private & 0.00 & 0.00 & 25.06 & 100.00 & 12.53 & 50.00 \\
    All public & 74.94 & 100 & 0.00 & 0.00 & 37.47 & 50.00  \\
    \midrule
    MLP &  86.29 & 82.32 & 53.52 & 60.89 & 69.90 & 71.60 \\
    GPA$^\star$ & 75.30 & 97.62 & 37.25 & 4.22 & 56.28 & 50.92  \\
    GPA$^\diamond$ &  74.94 & 100 & 0.00 & 0.00 & 37.47 & 50.00  \\
    GIP$^\triangle$ & 74.94 & 100 & 0.00 & 0.00 & 37.47 & 50.00  \\
    GA-MLP &  88.87 & 77.71 & 51.53 & 70.89 & 70.20 & 74.30  \\
    \midrule
    Strategy-1 &  94.76 & 55.05 & 40.34 & 90.89 & 67.55 & 72.97  \\
    Strategy-2 &  89.67 & 73.55 & 48.55 & 74.67 & 69.11 & 74.11  \\
    \bottomrule \addlinespace[\belowrulesep]
    \multicolumn{7}{l}{\parbox{0.9\columnwidth}{\scriptsize{KEY -- P:~precision; R:~recall, BA:~Balanced accuracy; MLP:~multi-layer perceptron; GA:~graph-agnostic baseline;
    GPA:~Graph Privacy Advisor~\cite{Stoidis2022BigMM}; $\star$:~GPA adapted to the hybrid approach; $\diamond$:~adapted GPA with corrected implementation of adjacency matrix; $^\triangle$:~GIP~\cite{Yang2020PR} adapted to the hybrid approach, using cardinality and confidence as object features and privacy nodes with zero-initialised features.
    }}}\\
    \end{tabular}
    \label{tab:classres}
    \vspace{-12pt}
\end{table}

\section{Person-centric classification}

Based on the outcomes of the previous analysis, we devise two person-centric decision strategies that act directly on the objects extracted by the vision models and the corresponding features. The first, simple strategy classifies an image as private if at least one person is detected, $x_0^c \geq 1$, where $x_0^c$ is the cardinality feature and the object $c$ corresponds to \textit{person}. 
The second, simple strategy includes an additional constraint that limits the number of people localised in an image, i.e., $x_0^c > 0 \wedge x_0^c \leq 2$, where $\wedge$ is the logical AND operator.

We report the performance of these two strategies on the testing set of PrivacyAlert in Table~\ref{tab:classres}. The second strategy achieves performance comparable to GA-MLP and outperforms MLP, especially in terms of recall on the private class and balanced accuracy.
The first strategy has lower balanced accuracy than the second strategy but achieves a recall of 90.89\% in the private class denoting that most of the private images contain people. Nonetheless, this first strategy has many false positives (a precision of 40.43\% in the private class), indicating that images with people are not necessarily private. The more restrictive condition of the second strategy better balances the issues of the first strategy, but the recall for private images is limited to 74.67\%.

\section{Conclusion}

In this paper, we used post-hoc explainability to identify and quantify  objects contributing to the decision of image privacy classification models, which are trained on concepts extracted from an image by a pre-trained detector.
The explainability analysis showed that privacy models, such as MLP and GA-MLP, are biased towards the presence of the object \textit{person}. Based on this finding, we devised two simple person-centric strategies that achieve comparable overall classification performance to that of the state-of-the-art  models consodered in the comparison.

Future work will extend the explainability analysis to other publicly available datasets, such as VISPR~\cite{Orekondy2017ICCV}, IPD~\cite{Yang2020PR} and DIPA~\cite{Xu2023_DIPA}, and other models with different concepts and  features~\cite{Baranouskaya2023ICIP,Yang2020PR,Tonge2018AAAI,Stoidis2022BigMM}. We will also include and compare the results of other explainability methods~\cite{Ribeiro2016ICMLw_LIME,Ribeiro2016KDD_LIME,Petsiuk2018BMVC_rise,bilodeau2024impossibility}.

\appendix

\section*{Appendix}

We provide details of the methods considered for our evaluation and explainability analysis. Specifically, we provide the rationale behind the chosen methods, a  review of the main aspects of each method. We also provide  details of the parameter settings,  and training details, and implementation details in common to all privacy models for a fair comparison under the settings designed in the main paper.
\blfootnote{Alessio Xompero and Myriam Bontonou equally contributed. Myriam Bontonou is also affiliated with Inserm, France, and Jean-Michel Arbona with Univ Lyon and LBMC Lyon, France.}

\section{Privacy models}

\subsection{MLP}

Tonge et al.'s method~\cite{Tonge2016AAAI} uses a convolutional neural network, pre-trained on ImageNet~\cite{Deng2009CVPR_ImageNet}, for multi-label object recognition. The feature vector with the confidence of the 1,000 objects is converted into binary values by assigning 1 to the top-$k$ most confident classes and 0 to the all the other classes.  The binarised feature vector is used as input to a classifier trained to predict the privacy of an image. Support Vector Machine was used as classifier and $k$ was set to 10 in the study~\cite{Tonge2016AAAI,Tonge2018AAAI}. 
Baranouskaya and Cavallaro~\cite{Baranouskaya2023ICIP} defined different input features, such as person presence, person cardinality, outdoor scene, and sensitive features (e.g., violence), and evaluated both a logistic regression and an MLP as privacy models. 

Following the ideas and results of these two studies, we devise a baseline that aims to reproduce the method but using the objects and their features as defined in the main paper (see Section 2). Specifically, we replace the multi-label object recognition with the object detector, and the binary feature vector with the cardinality and confidence features. We also use an MLP as privacy classifier given its best-performing results in Baranouskaya and Cavallaro's work~\cite{Baranouskaya2023ICIP}. We simply refer to this baseline as MLP.

\subsection{Graph-based methods} 

GIP~\cite{Yang2020PR} and GPA~\cite{Stoidis2022BigMM} belong to the two-stage end-to-end training-based category. Both methods model a graph of objects and two additional nodes representing the public and private classes (privacy nodes). The 80 COCO categories~\cite{Lin2018ECCV_COCO} are used as objects. 

GIP relates objects and privacy nodes with a weighted, undirected, bipartite graph using the frequency of each object with respect to all images labelled as either private or public in a given dataset~\cite{Yang2020PR}. A convolutional neural network (VGG-16) is fine-tuned to extracts deep features from the regions of interest localised in an image and associated with the corresponding object node in the graph. The privacy nodes are initialised with the deep features extracted from the whole image by another fine-tuned convolutional neural network (ResNet-101). When objects are not localised in an image, their features are initialised to 0. All features are also complemented with a 1-hot encoding vector to distinguish the privacy nodes, the object nodes, and the object nodes with zero-initialised features. 

GPA relates objects with each other by finding at least one co-occurrence of the objects in the dataset, resulting in an unweighted and undirected graph. GPA uses cardinality as object features and initialises the features of the privacy nodes with the logits from a trainable fully connected layer that maps the outputs (logits) of a ResNet-50 pre-trained for scene recognition to the two privacy classes. The scene classifier is also fine-tuned during the training of GPA. Both GPA and GIP use a Graph Reasoning Model~\cite{Wang2018IJCAI} to propagate and refine the node features according to the modelled graph structures, and then use a fully connected layer for the final classification. The Graph Reasoning Model consists of three layers of Gated Graph Neural Network~\cite{Li2016ICLR_GGNN} and a modified Graph Attention Network~\cite{Velickovic2018ICLR_GAT,Wang2018IJCAI}. 

\subsection{From end-to-end to a hybrid approach} 

To adapt the two methods to a two-stage hybrid approach, we decoupled the graph component (Graph Reasoning Model and fully connected layer) from the CNNs, and we initialise the nodes with the cardinality and confidence features obtained from the pre-trained object detector. This means that there is no longer the end-to-end training of the whole pipeline and fine-tuning of the CNNs. 

Under our setting, we cannot initialise the privacy nodes of GIP with the high-dimensionality (4,096) feature vectors extracted by ResNet-101 and hence we initialise the features of the two nodes to 0. We refer to this model as GIP$^\triangle$ in Table 1 of the main paper. Note that GIP was trained and evaluated only on the Image Privacy dataset~\cite{Yang2020PR}, whereas we train a new GIP model trained only on PrivacyAlert. 

Similarly, we removed the dependency of the scene classifier and the trainable fully connected layer for GPA. Because of the presence of the privacy nodes, we also discard the background category that was included to account for images with no detected objects. We therefore train a model as close as possible to the original implementation\footnote{\url{https://github.com/smartcameras/GPA/}} where the features of the object nodes are the cardinality and the binary flag\footnote{In our experiments, we found that the flag does not provide any contribution to the model.}. However, we replace the features of the privacy nodes with pseudo-randomly generated values in the interval $[-20,20]$ according to the range of the logits estimated by the fine-tuned CNN to simulate a non-optimised and non-zero initialisation of the features.  We refer to this model as GPA$^\star$ in Table~1 of the main paper. Note that we also evaluated a variant with zero-initialisation of the features of the privacy nodes and we obtained the same results.

As we noticed a misplacement of the adjacency matrix in the original implementation, we also corrected this error and train a second model. For this second model, we use both cardinality and confidence features, without the binary flag, for a fair comparison with the other models. We refer to this model as GPA$^\diamond$ in Table~1 of the main paper. We also tried with either of the two features, as well as using the projection to a higher dimensionality as done for GA-MLP, but all of these models degenerate to predicting a single class.

\section{Parameters setting and training details}

\noindent{\bf Object detector.} We use YOLOv3~\cite{Redmon2018YOLOv3}, pre-trained on the 80 categories of COCO~\cite{Lin2018ECCV_COCO}, as object detector. When localising the objects, we allow a maximum of 50 objects for each image while retaining the most confident ones after re-ranking. We also use a minimum threshold of 0.6 and a non-maximum suppression threshold at 0.4. According to the detector settings, we resize images to a resolution of 416$\times$416 pixels. Note that these settings are different from GIP and GPA, which limit the maximum number of regions of interest only to 12. Moreover, GIP used Mask R-CNN~\cite{He2017ICCV_MaskRCNN} as object detector with a threshold of 0.7 on the object confidence and the weighted edges of their modelled graph included images from the testing set (data leakage). On the contrary, GPA used YOLOv3~\cite{Redmon2018YOLOv3} with a threshold of 0.8 on the object confidence. Our choice to decrease the threshold is to allow the localisation of more objects in an image, increasing the detected categories and the cardinality for more discriminative features. However, the lower threshold can also result in more false positives and affecting the input features of the privacy model that should be designed to handle noisy data.

\noindent{\bf Training.} For reproducibility of models and experiments, we set the seed to an arbitrary value of 789.  Note that we do not analyse variations in the performance due to multiple and different seeds, which is beyond the scope of this paper. As training strategy, we follow the recipe of Benchmarking Graph Neural Networks~\cite{Dwivedi2023JMLR}. We use Adam as optimizer~\cite{Kingma2015ICLR} with an initial learning rate of 0.001 and without weight decay. We schedule the learning rate to halve if the balanced accuracy of the validation set does not improve for at least 10 epochs (patience). We use early stopping to interrupt the training of the models if the learning rate decreases to a value lower than 0.00001 or the training time lasts longer than 12 hours. In case none of the two conditions is satisfied, we also set the maximum number of epochs to 1,000. Note that we save the model at the epoch with the highest balanced accuracy in the validation split and We use this model for the evaluation on the testing split. Moreover, we set the batch size to 100.

\section{Implementation}

We implement all models using PyTorch 1.13.1. We use the PyTorch Geometric library for GIP, GPA, and GA-MLP. We trained all models on a Linux-based machine with a NVIDIA GeForce GTX 1080 Ti (12 GB RAM). To ensure the fairness of the benchmark, all methods share the same training and testing software (i.e., only the model is replaced).

\section*{Acknowledgements}

\noindent This work was supported by the CHIST-ERA programme through the project GraphNEx, under UK EPSRC grant EP/V062107/1 and France ANR grant ANR-21-CHR4-0009.

{\small
\bibliographystyle{ieee_fullname}
\bibliography{main}
}

\end{document}